\documentclass[sigconf]{acmart} 

\renewcommand\footnotetextcopyrightpermission[1]{}
\AtBeginDocument{%
  }

\usepackage{hyperref}       
\usepackage{url}            
\usepackage{booktabs}       
\usepackage{amsfonts}       
\usepackage{nicefrac}       
\usepackage{microtype}      
\usepackage{xcolor}         
\definecolor{revisionpink}{RGB}{220,20,120} %

\colorlet{red}{black}
\colorlet{blue}{black}

\usepackage{tcolorbox}      

\usepackage{amsmath}
\usepackage{amsfonts}
\usepackage{amsthm}
\usepackage{multirow}

\usepackage{tikz}
\usepackage{pgfplots}
\pgfplotsset{compat=1.18} 
\usepackage{subcaption}

\usepackage[ruled,vlined,linesnumbered]{algorithm2e} 
\usepackage{multirow} 

\setcopyright{none}
\copyrightyear{2026}
\acmYear{2026}
\acmDOI{}
\acmConference[]{}{}{}

\begin{document}

\title{Inferring Causal Relations between Two Sequences of Events with Language Models}

\author{Nishchal Prasad}
\email{prasadnishchal.np@gmail.com}
\affiliation{
  \institution{INRIA, Univ. Grenoble Alpes, CNRS, LIG}
  \city{Grenoble} \country{France}
}

\author{Eric Gaussier}
\email{eric.gaussier@univ-grenoble-alpes.fr}
\affiliation{
  \institution{Univ. Grenoble Alpes, CNRS, Grenoble INP, LIG}
  \city{Grenoble} \country{France}
}

\author{Emilie Devijver}
\email{emilie.devijver@univ-grenoble-alpes.fr}
\affiliation{
  \institution{Univ. Grenoble Alpes, CNRS, Grenoble INP, LIG}
  \city{Grenoble} \country{France}
}

\author{Alexander Obeid Guzman}
\email{alexander.obeid-guzman@inria.fr}
\affiliation{
  \institution{Univ. Grenoble Alpes, CNRS, Inria, Grenoble INP, LIG}
  \city{Grenoble} \country{France}
}

\author{Armen Aghasaryan}
\email{armen.aghasaryan@nokia-bell-labs.com}
\affiliation{
  \institution{Nokia Bell Labs}
  \city{Paris-Saclay} \country{France}
}

\author{Gregor G\"ossler}
\email{gregor.goessler@inria.fr}
\affiliation{
  \institution{Univ. Grenoble Alpes, CNRS, Inria, Grenoble INP, LIG}
  \city{Grenoble} \country{France}
}

\renewcommand{\shortauthors}{Prasad et al.}

\renewcommand{\shortauthors}{Prasad et al.}

\begin{abstract}
Causal AI is a branch of Artificial Intelligence which helps understand and reason about cause and effect relationships, not just patterns or correlations. Causal discovery aims to infer elements of the underlying causal structure—often represented as a directed graph—from observational and, when available, interventional data. While causal discovery is the fundamental step for moving beyond mere associations toward genuine understanding, and thus the basic building block of causal AI, it becomes intrinsically difficult when causal relations must be inferred from single observations. In such situations, standard causal discovery methods cannot be used and one has to identify causal relations from limited amount of information. This is typically the case for, \textit{e.g.}, sequences of events produced by different alarms which need to be analyzed on the fly to detect abnormal phenomena, which are usually rare. We show in this study that it is possible to leverage the predictive power of Large Language Models (LLMs) to infer causal relations between only two sequences of events. This approach, which is validated on both synthetic and real data, provides better results than standard causal discovery algorithms on several time series data, even though these data were converted into smaller, single observed sequences.
\end{abstract}




\maketitle
\section{Introduction}

Inferring the direction of causation between two co-occurring processes is a foundational challenge across scientific and engineering disciplines. In modern telecommunications networks, for instance, hundreds of devices continuously emit discrete alarms. Correctly orienting the causal direction, determining whether an anomaly in a radio module caused a failure in the core network, or vice versa, is what separates a useful root-cause diagnosis from a mere observation of correlation \cite{zhang2021influence, shi2025causal}. Similar challenges arise in IT monitoring, where metric-derived events trigger cascading incidents \cite{aitbachir2023itmonitoring, zanCIKM}, and in medical workflows, where a patient's symptom progression and subsequent clinical interventions form complex, interleaved event streams \cite{cornanguer2025niagara}. 

The standard formulation of this task is \emph{inferring pairwise causal direction}: given observations of two correlated variables $X$ and $Y$, determine the causal direction ( $\!\to\!$ ) i.e. whether $X \!\to\! Y$ or $Y \!\to\! X$. Since conditional independence tests  require a third variable to be informative, constraint-based methods fail in this setting. Thus, existing approaches for pairwise data rely on exploiting structural asymmetries in the data-generating mechanism \cite{mooij2016distinguishing}. For continuous variables with large, independent and identically distributed (i.i.d.) samples, a rich literature has successfully developed criteria based on additive-noise models \cite{hoyer2009nonlinear}, post-nonlinear models \cite{zhang2009postnonlinear}, and conditional divergence \cite{duong2022cdci}. 

The landscape changes when $X$ and $Y$ are temporal \emph{event sequences} representing unobserved processes, and we are given only a \emph{single observation} of each. 
%
 Consider a network operator who observes a single trace of events on module $X$ and a single trace on module $Y$ during one outage, or a clinician who sees one patient's symptom timeline and one treatment timeline during a single hospital stay, or a biologist confronted with one transcriptional pulse pair from a single experiment of tracking two different genes (gene $X$ and gene $Y$) simultaneously. Here no population exists from which additional samples can be drawn; the observation is, by construction, of size one. Existing pairwise causal direction discovery methods, regardless of whether they target continuous data \cite{hoyer2009nonlinear,janzing2012igci,marx2017slope} or event sequences with many recurrences \cite{budhathoki2018cute,cuppers2024cascade,cornanguer2025niagara,xu2016hawkes,qiao2023shp}, assume access to repeated observations to either estimate distributions, or fit functional models, or aggregate Hawkes-style intensities. 
 \textbf{
 Prior to this work, the problem of orienting the causal direction between two sequence-based variables from a single observation of each remained unaddressed in the literature.} 

To close this gap, we propose a general inference framework for pairwise causal direction discovery in single observations based on 
the zero-shot sequence modeling capabilities of LLMs. Rather than relying on statistical sampling or temporal ordering, we treat this problem 
as a problem of evaluating 
the asymmetry of sequence generating probabilities. If $X$ causes $Y$, the sequence generation process dictates that the joint probability factorized in the true causal direction, where $Y$ is structurally conditioned on $X$, is higher than the anti-causal factorization. We capture this through a scoring framework that evaluates the likelihood of each sequence of events.

To compute these probabilities without training data, we empirically leverage pre-trained Large Language Models (LLMs) purely as zero-shot 
probability density estimators. Recent findings demonstrate that modern LLMs are highly capable "general pattern machines" \cite{mirchandani2023patternmachines}. Because of the in-context learning mechanisms through induction heads \cite{induction-head-in-context-1}, LLMs can evaluate the structural dynamics of arbitrary, out-of-distribution sequences without requiring fine-tuning \cite{gruver2024large}.


In this work, we propose a novel, observation-driven inference framework for pairwise causal discovery from a \emph{single run} of a system. Instead of querying an LLM for semantic knowledge, we encode the event symbols and utilize the LLM as a probability density estimator to obtain the likelihood of a hypothesized sequence generation process. Our main contributions are the following:
\begin{itemize}
    \item We formalize the single observation causal discovery problem for event sequences and introduce a \emph{Sequence Generation Rule}. This rule infers causation by comparing the auto-regressive negative log-likelihoods (NLL) of the sequences under different causal hypotheses.
    \item We use pre-trained LLMs as domain-agnostic probability density estimators. To mitigate the token's semantic bias and adapt an LLM to out-of-distribution event sequences, we introduce two practical mechanisms: \emph{random encoding mapping} to get expected likelihoods across arbitrary sequence encoding, and \emph{sequence replication} for better in-context learning.
    \item We rigorously develop and evaluate our framework on a synthetic data; varying sequence lengths, encoding configurations, and LLM backbones, and test on real-world IT-monitoring data. We demonstrate that our approach reliably discovers the causal direction from single sequence pairs and significantly outperforms traditional baselines. 
\end{itemize}

To the best of our knowledge, this is the first method for orienting causation between two event sequences from a single observation, and to leverage LLMs not as semantic knowledge bases but as structural probability estimators for causal inference. 

The remainder of the paper is organized as follows.
Section~\ref{sec:sota} reviews related work. 
Section~\ref{sec:methodology} introduces our proposed method: first detailing the LLM-based estimation in Section \ref{sec:llm_precedence_estimation} and detailing practical considerations in Section \ref{sec:practical_consideration}, leading to the causal discovery method developed in Section \ref{sec:algorithm}. 
Section~\ref{sec:experimental_setup} describes the experimental setup, and Section~\ref{sec:results} presents our results, on synthetic data (Section~\ref{sec:results:synthetic}) and on real-world IT monitoring data (Section ~\ref{sec:results:real}).
Finally, Section~\ref{sec:conclusion} concludes the paper with Section~\ref{sec:limitations} describing some limitations of the work.


%

%




\section{Related Work}
\label{sec:sota}
We look at the prior work from the perspective of the 
data on which causal direction is inferred: continuous real-valued data (multi-observation), event sequence (multi observation, single observation), and close with the line of work that uses LLMs for causal discovery. In this section, $X$ and $Y$ are the two variables whose causal relationship is being studied.

\paragraph{{Pairwise causal direction from continuous data:}} 

In the classical setting, we observe $n$ i.i.d.\ draws $\{(X^{(i)}, Y^{(i)})\}_{i=1}^{n}$ of variables $X \in \mathbb{R}^{p}$ and $Y \in \mathbb{R}^{q}$ (where $p,q \geq 1$). Two main types of methods dominate this area.

The first type, \emph{functional causal models}, assumes a parametric form for the causal mechanism and exploits the resulting statistical asymmetries. For instance, Linear Non-Gaussian Acyclic Models (LiNGAM)\cite{shimizu2006lingam}, Additive Noise Models \cite{hoyer2009nonlinear}, and Post-Nonlinear Models \cite{zhang2009postnonlinear} infer causation 
by identifying the unique direction in which the estimated noise is independent of the presumed cause. Alternatively, regression error-based method 
\cite{bloebaum2019reci} infers the causal direction by comparing the mean-squared errors of a regression model's fit in both directions.

The second type, \emph{independence-of-mechanism 
}, exploits the postulate that the marginal distribution $P(X)$ and the conditional mechanism $P(Y\mid X)$ are independent in the true causal direction. Within this type, Information-Geometric Causal Inference \cite{janzing2012igci} relies on orthogonality in information space. Conditional Distribution Similarity \cite{fonollosa2019cds} measures the  variability in the shape of $P(Y\mid X{=}x)$ across $x$. And Conditional Divergence-based Causal Inference \cite{duong2022cdci} captures a similar intuition via the average divergence between normalised conditionals. Algorithmic information variants build on this same principle without assuming a fixed functional form. They use Minimum Description Length (MDL) to approximate the Kolmogorov complexity $K(P(X)) + K(P(Y\mid X))$
\cite{marx2017slope,marx2018crack,marx2021formally}.

Beyond these two primary types, \emph{learning-based approaches} such as RCC/NCC \cite{lopezpaz2015rcc,lopezpaz2017discovering} and Meta-CGNN \cite{ton2021meta} re-frame causal discovery as a classification problem, leveraging kernel embeddings or neural networks to extract features from $P(X,Y)$.

Most of these methods are evaluated against the de-facto benchmark of 100 cause-effect pairs curated by Mooij et al. \cite{mooij2016distinguishing}. 
However, none of these approaches are directly applicable to our setting as they fundamentally rely on having a large number of i.i.d.\ draws from the joint distribution $P(X,Y)$.

\paragraph{{Pairwise causal direction from discrete event-sequence data:}}
A second class of methods takes $X \in \Sigma_x^{n_x}$ and $Y \in \Sigma_y^{n_y}$ with $n_x, n_y \geq 1$ discrete-valued sequences oriented by time, where $\Sigma_x$ and $\Sigma_y$ are the possible discrete value sets. Within this class, the dependence on having repeated observations is even more central, and the literature splits along that axis.

In scenarios with multiple observations, i.e., multiple  sequences per pair of random variables, most existing approaches for event sequences assumes access to many concurrent or repeated sequences to estimate causal intensities or distributions. \emph{Granger-causal} approaches (such as transfer entropy \cite{schreiber2000transferentropy} and its extensions \cite{shojaie2022granger}) test whether the past of $X$ improves the prediction of $Y$ beyond $Y$'s own past, requiring sufficiently long sequences for reliable conditional entropy estimates. \emph{Hawkes-process} methods \cite{xu2016hawkes,zhou2013adm4,achab2018nphc,zhang2020cause,jalaldoust2022mdlh,cai2022thps,qiao2023shp} model how events trigger one another, but similarly depend on observing enough events to fit their underlying functions. 
A separate \emph{algorithmic} line treats event sequences through compression. For example, CUTE \cite{budhathoki2018cute} infers direction by comparing sequential normalised maximum-likelihood code lengths $\ell(y^n) - \ell(y^n \mid x^n)$ and vice versa. Origo \cite{budhathoki2018origo} does the same for binary tabular data through MDL with decision-tree compressors. Most recently, CASCADE \cite{cuppers2024cascade} learns a one-to-one cause-effect matching by combining MDL with the algorithmic Markov condition, and NIAGARA \cite{cornanguer2025niagara} extends this to interval-based events with rich parent interactions. Another line of \emph{Logic-based} frameworks, beginning with Kleinberg and Mishra \cite{kleinberg2009temporal,kleinberg2013rare}, encode causal hypotheses as PCTL (Probabilistic Computation Tree Logic) and aggregate evidence across many event traces. Common to all of the above, \emph{multiple observations}; i.e. either many sequence pairs or many recurrences of each event type;  
are essential.

When considering a single observation (only one $X$ and one $Y$), 
there is, to our knowledge, \emph{no prior work} for inferring causal relation with no extra population data to draw from. The closest theoretical foundation is the algorithmic Markov condition itself \cite{janzing2010algorithmic}, which establishes that causal inference is in principle possible from single objects by comparing Kolmogorov complexities of the two factorisations of the joint distribution. 
However it leaves open the question of a practical, computable instantiation when the objects are event sequences and no domain-specific compressor is available. Existing MDL or compression-based pairwise inference methods (Origo \cite{budhathoki2018origo}, CUTE \cite{budhathoki2018cute}, Slope \cite{marx2017slope}, the stochastic-complexity framework of \cite{budhathoki2017stochastic}) all rely on compressors 
that are themselves trained or fit on the input sequence and require enough recurrences within that sequence to compress it meaningfully. In the single observation setting, those compressors degenerate, motivating our use of a \emph{general-purpose, pre-trained} LLM as a parameter-free probability density estimator.

\paragraph{LLMs for causal discovery:}

The intersection of LLMs and causal inference has recently garnered significant attention, though the vast majority of approaches rely on the models' internal trained knowledge. For example, Kiciman et al. \cite{kiciman2024causal} demonstrated that language models (LMs) achieve state-of-the-art performance in pairwise causal discovery by evaluating the semantic variable names (e.g., querying the model whether "Altitude" causes "Temperature"). 
Similarly, other works have used LLMs to extract causal graphs directly from text corpora by ``prompting'' the LLMs to identify causal relationships between textual entities \cite{long2024largelanguagemodelsbuild}. 
While effective, these approaches are strictly metadata-driven and rely entirely on the model having encountered the causal mechanism during pre-training, they are ineffective on the event sequences whose semantic content is, by design, absent.

In contrast to prior work, we focus on inferring causal direction from a single observation of event sequence, leveraging LLMs not as semantic knowledge bases but as structural probability density estimators. This paradigm aligns with recent evidence of LLMs' zero-shot sequence modeling capabilities. 
For example, Gruver et al. \cite{gruver2024large} showed that LLMs can predict out-of-distribution time series data directly by recognizing abstract numerical patterns in-context, and Mirchandani et al. \cite{mirchandani2023patternmachines} show that the pattern completion ability persists even when sequences are expressed in tokens randomly drawn from the vocabulary. Similarly, work on induction heads \cite{induction-head-in-context-0, induction-head-in-context-1} also shows evidences on how transformers copy and extrapolate complex, non-linguistic patterns without additional training. Alternatively, LLMs are also able to compress out-of-distribution data better than traditional compressors \cite{deletang2024language}. 

\section{Methodology}
\label{sec:methodology}
We consider pairs of observed sequences whose underlying generation processes are driven by two independent mechanisms: one governing a cause process, and another process generating the effect from the cause. Both causes and effects are \textit{unobserved processes}\footnote{Processes that are not directly visible but are detectable through their effects.}. Let $\mathcal{P}_{h}$ denote the space of all possible unobserved/hidden process. We denote the unobserved/hidden process by a $^{*}$, such as $X^{*}$ or $Y^{*} \in \mathcal{P}_{h}$. In practice, one only observes two sequences of symbols, $X^{(0)}$ and $Y^{(0)}$ corresponding to two causally related unobserved process $X^{*}$ and $Y^{*}$, where the vocabulary of symbols, hereafter called \textit{encoding}, for $X^{(0)}$ and $Y^{(0)}$ has been arbitrarily chosen\footnote{The notation $^{(0)}$ serves as a reminder that the encoding is arbitrary: it is just one possible encoding, numbered $0$, among many other possible ones.}. 
Our goal in this study is to determine, from $X^{(0)}$ and $Y^{(0)}$, whether $X^{*}$ causes $Y^{*}$ or whether $Y^{*}$ causes $X^{*}$. The problem of ``inferring causal direction between two causally related event sequences from a single observation'' is defined below.


\begin{tcolorbox}[
    boxrule=1pt,
    colback=white!95!black,
    sharp corners,
    left=2.5pt,right=2.5pt,top=2.5pt,bottom=2.5pt,
    boxsep=1pt
]
\textbf{Problem Definition.}\label{problem-definition} Given only one observation $X^{(0)}$, $Y^{(0)}$ for $X^{*}$, $Y^{*}$, and assuming that a direct causal relation exists between them, determine whether $X^{*}$ causes $Y^{*}$ or $Y^{*}$ causes $X^{*}$ (i.e. $X^{*} \rightarrow Y^{*}$ or $Y^{*} \rightarrow X^{*}$).  
\end{tcolorbox}

To model causal relationships, we introduce two mechanisms. The first one, a cause-generating mechanism, is characterized by the probability $P_{C}$ that a process is a cause ($C$). Similarly, the second mechanism is characterized  by the conditional probability $P_{E|C}$ that a process is the effect ($E$) of a given cause ($C$). We then define the causal probability as: 
\begin{alignat}{3}
    P_{CE}: ~& \mathcal{P}_{h} \times \mathcal{P}_{h} & \, \longrightarrow & \, [0,1] \nonumber \\
    & (X^{*},Y^{*}) & \, \longrightarrow & \, P_{C}(X^{*}) \times P_{E|C}(Y^{*}|X^{*}).\nonumber
\end{alignat}
It is a non-symmetric probability distribution over $\mathcal{P}_{h} \times \mathcal{P}_{h}$. It is important to note that $P_{CE}$ is not a joint probability distribution over the cartesian product space $\mathcal{P}_{h} \times \mathcal{P}_{h}$ as it does not provide the probability that $X^{*}$ and $Y^{*}$ simultaneously co-occur, but rather the probability that $X^{*}$ is a cause of $Y^{*}$, based on $P_C$ and $P_{E|C}$. Indeed, one has:
\begin{equation}
\left\{ \begin{array}{ll} 
    \sum_{Y^{*}} P_{CE}(X^{*},Y^{*}) & = P_{C}(X^{*}), \nonumber \\
    \sum_{X^{*}} P_{CE}(X^{*},Y^{*}) & = \sum_{X^{*}} P_{C}(X^{*}) P_{E|C}(Y^{*}|X^{*}) \nonumber 
    \end{array} \right. 
\end{equation}
This interpretation motivates the following rule for inferring causal direction between pairs of processes:
\begin{tcolorbox}[
    boxrule=1pt,
    colback=white!95!black,
    sharp corners,
    left=2.5pt,right=2.5pt,top=2.5pt,bottom=2.5pt,
    boxsep=1pt
]
\textbf{Rule A:} \label{Rule-A} The causal direction between 
$X^{*}$ and $Y^{*}$, assuming that a direct causal relation exists between them, is given by the highest $P_{CE}$, that is: if $P_{CE}(X^{*},Y^{*}) > P_{CE}(Y^{*},X^{*})$, then $X^{*}$ causes $Y^{*}$; $Y^{*}$ causes $X^{*}$ otherwise.
\end{tcolorbox}
%
\noindent This rule reflects the inherent asymmetry of the data generation process rather than mere temporal precedence; specifically, the factorization corresponding to the true causal direction naturally yields a higher overall generation probability than the reverse direction.
This is also reflected in the asymmetry of $P_{CE}$. In addition, we define the causal score $S_{X^{*} \rightarrow Y^{*}}$ as:
\[
S_{X^{*} \rightarrow Y^{*}} \triangleq - \log P_{CE}(X^{*},Y^{*});
\]
and we denote
\begin{align*}
S_{C}(X^{*}) &= - \log P_{C}(X^{*})\\
S_{E|C}(Y^{*}|X^{*}) &= - \log P_{E|C}(Y^{*}|X^{*}).
\end{align*}
Then, Rule A is equivalent to selecting the causal direction yielding the lowest causal score between $S_{X^{*} \rightarrow Y^{*}}$ and $S_{Y^{*} \rightarrow X^{*}}$.

\subsection{LLM-based estimation}
\label{sec:llm_precedence_estimation}
As  mentioned before, one does not directly observe the hidden process $X^{*}$ and $Y^{*}$ but rather two sequences of symbols, $X^{(0)}$ and $Y^{(0)}$, generated by them. To infer causal relationships, we leverage LLMs to estimate the probabilities $P_C$ and $P_{C|E}$ over the unobserved processes from the observed sequences. This choice is motivated by LLMs' demonstrated ability to model sequence likelihoods with state-of-the-art performance across diverse tasks, from language modeling to pattern recognition~\cite{Survey_LLMcapabilities}. 

During pre-training, LLMs are optimized to minimize the negative log-likelihood (NLL) of the next predicted token, which is mathematically equivalent to minimizing cross-entropy loss. As a result, these models are explicitly trained to estimate the exact quantities required by our scoring framework ($S_{\cdot \rightarrow \cdot}$), which is the joint probability of a sequence of discrete symbols, factorized auto-regressively as $P(x_1, \dots, x_n) = \prod_{i=1}^{n} P(x_i \mid x_{<i})$. This formulation enables LLMs to naturally assign probabilities ({$\widehat {P_{\text{llm}}}$}) to entire sequences by estimating the conditional probability of each token given its preceding context. 
 
Furthermore, to avoid dependence on a given arbitrary encoding of the sequences, we consider expectations over all possible encodings. Denote by  $\mathcal{E}^{X}$ (resp. $\mathcal{E}^{Y}$) the set of alternative encodings (see Section~\ref{subsec:encoding}) of $X^{(0)}$ (resp. $Y^{(0)}$), and, for any $e \in \mathcal{E}^{X}$ and $e' \in \mathcal{E}^{Y}$, by $X^{(e)}$ (resp. $Y^{(e')}$) the sequence obtained from $X^{(0)}$ (resp. $Y^{(0)}$) by applying encoding $e$ (resp. $e'$).
We  can either estimate the likelihood of $X^{*}$ and $Y^{*}$ over all possible encodings by:
%
%
\begin{equation}
\label{eq:expectation_over_encodings}
\left\{ \begin{array}{lll}
    \hat{P}_{C}(X^{*}) & = \mathbb{E}_{e \in \mathcal{E}^{X}} [ \widehat{P_{\text{llm}}}(X^{(e)})], \\
    \hat{P}_{E|C}(Y^{*}|X^{*}) & = \mathbb{E}_{\substack{e \in \mathcal{E}^{X} \\ e' \in \mathcal{E}^{Y}}} [ \widehat{P_{\text{llm}}}(Y^{(e')}|X^{(e)})],
    \end{array} \right.
\end{equation}
or estimate the causal scores by:
\begin{equation}
\label{eq:expectation_over_scores}
    \hat{S}_{X^{*} \rightarrow Y^{*}}  = \mathbb{E}_{\substack{e \in \mathcal{E}^{X} \\ e' \in \mathcal{E}^{Y}}} [- \log  \widehat{P_{\text{llm}}}(X^{(e)}) - \log  \widehat{P_{\text{llm}}}(Y^{(e')}|X^{(e)})]. 
\end{equation}
%
%
\noindent In practice, both approaches lead to estimated causal scores $\hat{S}_{X^* \rightarrow Y^*}$ and $\hat{S}_{Y^* \rightarrow X^*}$, for $S_{X^* \rightarrow Y^*}$ and $S_{Y^* \rightarrow X^*}$ respectively, which, even if not equivalent, yield very similar results in practice. In the remainder, we thus focus on the estimates provided by Eq.~\ref{eq:expectation_over_scores}.

Substituting these estimates into Rule A gives us the \textit{sequence generation} rule which aims to compare the probability of generating sequences for $Y^{*}$ \textit{from} sequences for $X^{*}$ against the probability of generating sequences for $X^{*}$ \textit{from} sequences for $Y^{*}$.
\begin{tcolorbox}[
    boxrule=1pt,
    colback=white!95!black,
    sharp corners,
    left=2.5pt,right=2.5pt,top=2.5pt,bottom=2.5pt,
    boxsep=1pt
]
\textbf{Sequence Generation Rule:} Given $X^{*}$ and $Y^{*}$ and assuming that a direct causal relation exists between them, the orientation of the causal direction is given by the smallest $\hat{S}_{. \rightarrow .}$: if $\hat{S}_{X^{*} \rightarrow Y^{*}} < \hat{S}_{Y^{*} \rightarrow X^{*}}$, then $X^{*}$ causes $Y^{*}$; $Y^{*}$ causes $X^{*}$ otherwise.
\end{tcolorbox}
%



\subsection{Practical considerations}
\label{sec:practical_consideration}
Relying completely on the assumption that an LLM has been trained on enough examples resembling the probability distribution of the event sequence being tested can be too strong. Moreover, since we restrict ourselves to single observations, we cannot train or fine-tune the LLM on a set of such sequences to exploit them later during inference. This constraint forces us to look for alternatives to align the predictive power of a pre-trained language model to our out-of-distribution sequences. To do so, we adapt two main practical considerations given the nature of LLMs: (i) token mappings through 'encoding' to remove the effect of LLM's semantic bias for one specific encoding, and (ii) 'replication' to increase the length of shorter sequences and improve in-context learning.


\subsubsection{{Encodings}}
\label{subsec:encoding}

LLM-based probability estimates for a given sequence are biased by token choice, as their training objectives and data are primarily aligned with linguistic semantics. 
 For event sequences, where each event sequence can have its own probability distribution and structure that are out-of-distribution of an LLM's training data, the semantics of the sequence interpreted by the LLM can influence the probability estimates heavily. 
 
To mitigate this bias, we introduce random token encodings.
Concretely, to avoid the dependence on a given encoding of a sequence $X$, we generate multiple encodings by randomly mapping its distinct symbols to a randomly selected subset of a universal vocabulary via bijective transformations. 
This is done as follows. 

Let $ X = (x_1, x_2, \dots, x_n) $ be a finite observed sequence of symbols, and let
$\Sigma = \{\, d \mid d \text{ appears in } X \,\} $ denote the alphabet size (set of distinct symbols occurring in $X$), with $ |\Sigma| = m $. Let $U$ be a universal set of symbols such that $|U| \ge m $. For each encoding, we construct a target alphabet $ \Sigma' \subseteq U $ such that $ |\Sigma'| = m $. This subset $ \Sigma' $ is chosen (here uniformly at random) from all possible $ m $-element subsets of \( U \). The encoding function $e$ is defined as a bijection, 
\[
e : \Sigma \to \Sigma' .
\]
Since $e$ is bijective, it is a re-labeling of the sequence $X$, producing a new observed sequence $X^{(e)} = (x_1^{(e)}, x_2^{(e)}, \dots, x_n^{(e)})$.

We then define the set of all encodings of $X$ as
\[
\mathcal{E}^X = \{\, X^{(e)} \mid e : \Sigma \to \Sigma',\ \Sigma' \subseteq U,\ |\Sigma'| = m,\ e \text{ bijective} \,\}.
\]
\noindent To generate elements of $ \mathcal{E}^X $, the following procedure is repeated:
\begin{enumerate}
    \item Sample a subset \( \Sigma' \subseteq U \) with \( |\Sigma'| = |\Sigma| \).
    \item Sample a bijection \( e : \Sigma \to \Sigma' \).
    \item Apply $e$ to $X$ to obtain the encoded sequence $X^{(e)}$.
\end{enumerate}
\noindent This is iterated $K = |\mathcal{E}^X|$ times to get $K$ encoded realizations of $X$. 
The random bijective mapping sets can always be made disjoint between a given sequence pair $X^*$, $Y^*$. 

Since $|\mathcal{E}^X| = \binom{|U|}{m} \cdot m!$, there are \( \binom{|U|}{m} \) choices for \( \Sigma' \) and \( m! \) bijections from \( \Sigma \) to \( \Sigma' \), which is too large for an exhaustive exploration. We therefore 
restrict both the size of $U$ and the number of bijections $K$ considered in practice (see Section \ref{sec:experimental_setup:model} for our design of experiments). 

\subsubsection{{Replication}}
\label{subsec:replication}

Given that shorter sequences can impact the in-context learning of the LLM and since the LLMs are strictly causal, 
we develop the concept of "replication" to artificially increase the length of shorter sequences and also incorporate pseudo-bi-directionality, which helps activate the induction heads to identify patterns in the sequences without explicit training. 

We rely on the in-context learning \cite{induction-head-in-context-1} capability of an LLM during inference to estimate the sequence probabilities, allowing it to adapt to patterns within the input sequence without additional training. 
A key limitation, however, of standard LLMs is their strict auto-regressive (left-to-right) causal attention, which constrains the model to only look at a sequence once. 
This unidirectional view can however fail to capture patterns that would be more apparent under a bidirectional lens. To address this, 
we 
use sequence ``replication''. 
Replication duplicates the sequence once, which helps the auto-regressive attention in LLMs to attend to the entirety of the original sequence when estimating the  probability of the replicated sequence. 
Given a finite sequence, 
$X=(x_1, x_2, \dots, x_n)$, we define the replicated context $X^{(r)}$ for a replication factor $r \in \{0, 1\}$ as:
\[ 
    X^{(r)} = \begin{cases} 
    X & \text{if } r = 0 \\
    X \oplus X & \text{if } r = 1 
    \end{cases}
\]
where $\oplus$ denotes concatenation. 
Using this, when $r=1$, a $j$-th token $x_j$ in the subsequent block $X$ is now estimated using the modified conditional probability $P(x_j \mid X^{(r-1)} \oplus x_{<j})$. Additionally, replication strengthens the LLM’s in-context learning ability by enhancing its induction-head  \cite{induction-head-in-context-0,induction-head-in-context-1} activation 
by increasing the frequency of latent patterns in a sequence. 

Evidence of different forms of replication (which is also referred to as ``repetition") can be found in improving embedding quality (\citet{repetition_0}), reasoning (\citet{read_twice_1}) and information recall (\citet{read_twice_0}) in standard auto-regressive LLMs.

\subsection{{LLM-based Causal Discovery Algorithm}}
\label{sec:algorithm}

Combining the theoretical foundation of the \textit{Sequence Generation Rule} (Section \ref{sec:llm_precedence_estimation}) and the practical mechanisms of random encodings and context replication (Section \ref{sec:practical_consideration}), we formulate our overall causal direction discovery method in Algorithm \ref{alg:causal_discovery}. 


For an observed sequence $X^{(0)}$ from $X^*$ 
of length $n$, let $X^{(e_k, r)}$ 
denote the sequence obtained after applying the encoding $e_k \in \mathcal{E}^X$ and applying the replication factor $r \in \{0, 1\}$ to $X^{(0)}$. Its marginal causal score for being the cause is: 
\begin{equation}
    S_{C}(X^{(e_k, r)}) = -\log \widehat {P_{\text{llm}}}(X^{(e_k, r)}) 
\end{equation}

Similarly, for an observed target sequence $Y^{(0)}$ from $Y^*$ of length $m$, let $Y^{(e_k', r)}$ 
be its encoded ($e_k' \in \mathcal{E}^Y$) and replicated ($r \in \{0, 1\}$) counterpart. The conditional causal score of $Y^{(e_k', r)}$ being the effect given the prior context sequence $X^{(e_k, r)}$ as the cause is given by: 
\begin{equation}
    S_{E|C}(Y^{(e_k', r)} \mid X^{(e_k, r)}) = -\log  \widehat {P_{\text{llm}}}(Y^{(e_k', r)} \mid X^{(e_k, r)}) 
\end{equation}


Algorithm \ref{alg:causal_discovery} outlines the batched procedure. \textcolor{black}{For each encoding $k$, $\mathcal{H}_{X\to Y}^{(k)}$ and $\mathcal{H}_{Y\to X}^{(k)}$ in Algorithm~\ref{alg:causal_discovery} (line 6,7) are the summands of Eq.~\eqref{eq:expectation_over_scores} under the two orientation hypothesis.}

For an observed batch of $X^{(0)}$, $Y^{(0)}$ pairs, we accumulate $S_{C}$ and $S_{E|C}$ (step $7 ~\&~ 8$) under both causal hypotheses ($X^* \rightarrow Y^*$ and $Y^* \rightarrow X^*$) in $H_{X^* \to Y^*}$ and $H_{Y^* \to X^*}$ across all $K$ encodings (step $9 ~\&~ 10$).

\begin{algorithm}[htbp]
\small
\SetAlgoLined
\KwIn{Observed Sequence Batch $(X^{(0)}, Y^{(0)}) \in \mathcal{D}$, Universal Set $U$, Encoding Size $K$, Replication factor $r \in \{0, 1\}$}
\KwOut{Inferred Causal Direction}

\ForEach{Batch $(X^{(0)}, Y^{(0)}) \in \mathcal{D}$}{\tcp{Initialize accumulators }
    $H_{X^* \to Y^*} \leftarrow 0, H_{Y^* \to X^*} \leftarrow 0$\; 
    
    \For{$k \leftarrow 1$ \KwTo $K$}{
        \tcp{1. Generate Bijective Mappings}
        Sample target alphabet $\Sigma' \subseteq U$ \\
        Sample bijections $e_k : \Sigma \to \Sigma', e_k' : \Sigma \to \Sigma'$\;
        Apply encoding: $X^{(e_k)} \leftarrow e_k(X^{(0)})$; $Y^{(e_k')} \leftarrow e_k'(Y^{(0)})$\;
        
        \tcp{2. Accumulate all $S$ scores (Replication)}
        \tcp{Hypothesis $X \to Y$}
        $\mathcal{H}_{X\to Y}^{(k)} \leftarrow S_{C}(X^{(e_k, r)}) + S_{E|C}(Y^{(e_k', r)} \mid X^{(e_k, r)})$\;
        \tcp{Hypothesis $Y \to X$}
        $\mathcal{H}_{Y\to X}^{(k)} \leftarrow S_{C}(Y^{(e_k', r)}) + S_{E|C}(X^{(e_k, r)} \mid Y^{(e_k', r)})$\;
        
        $H_{X^* \to Y^*} \leftarrow H_{X^* \to Y^*} + \mathcal{H}_{X\to Y}^{(k)}$\;
        $H_{Y^* \to X^*} \leftarrow H_{Y^* \to X^*} + \mathcal{H}_{Y\to X}^{(k)}$\;
    }
    
    \tcp{3. Final causal scores (Expectation over encodings, Eq. \eqref{eq:expectation_over_scores})}
    $\hat{S}_{X^* \to Y^*} \leftarrow H_{X^* \to Y^*} / K$\;
    $\hat{S}_{Y^* \to X^*} \leftarrow H_{Y^* \to X^*} / K$\;
    
    \tcp{4. Decision (Sequence Generation Rule)}
    \uIf{$\hat{S}_{X^* \to Y^*} < \hat{S}_{Y^* \to X^*}$}{
         $X^* \to Y^*$\;
    }
    \Else{
         $Y^* \to X^*$\;
    }
}
\caption{LLM-based Causal Direction Discovery Algorithm}
\label{alg:causal_discovery}
\end{algorithm}

As seen in Line 12 of Algorithm \ref{alg:causal_discovery}, the final score $\hat{S}_{. \rightarrow .}$ is obtained by averaging the scores over the $K$ encodings to approximate the expectation. One alternative to get the final orientation, apart from averaging, can be through max-voting amongst the discretely inferred directions of each encoding iteration. We performed experiments with this alternative, and it showed worse performance compared to averaging. We believe that this is because averaging takes into account the continuous causal scores of individual encodings, which naturally smoothens and separates away the token sensitivity in the final decision rule. In contrast, in voting, each individual vote rigidly includes a token-biased decision per encoding without accounting for the magnitude of the score difference. 



\section{Experimental Setup}
\label{sec:experimental_setup}
This section presents the 
hyper-parameters of our method (Sec.~\ref{sec:experimental_setup:model}), the test data (Sec.~\ref{sec:datasets}), and the baseline methods (Sec.~\ref{sec:baselines}).

\subsection{Hyper-parameters}
\label{sec:experimental_setup:model}
Since Algorithm \ref{alg:causal_discovery} can be computationally complex, given the size of the chosen language model, we fixed a moderately sized language model for our experiments. We mainly experiment with Llama-3.2-3B \footnote{\url{https://huggingface.co/meta-Llama/Llama-3.2-3B}}\cite{Llama3} as our main backbone language model based on its small size and state-of-the-art performance in the NLP benchmarks such as MMLU \cite{MMLU} and ARC-Challenge \cite{ARC_challenge}. 
In order to test our method's adaptability and variance with the choice of the LLM we also experimented with smaller LLMs (GPT-2\footnote{\url{https://huggingface.co/openai-community/gpt2}} \cite{gpt2} with $\sim$124M parameters and  Llama-3.2-1B \footnote{\url{https://huggingface.co/meta-Llama/Llama-3.2-1B}}\cite{Llama3} with 1B parameters). We conduct our experiments using one A100/H100 GPU, 
and to reduce GPU RAM overload the Llama variants are loaded in 16-bit floating point precision.

As discussed in Section \ref{sec:practical_consideration}, for a given  sequence $X$, the total number of possible encoding $K = |\mathcal{E}^X|$ is too large for exploration. Hence we choose to experiment with encodings $K$ $\in \{0, 1, 51, 311, 1111\}$. The choice of $K$ was random and motivated by having an odd (except $0$) and increasing integer. $K = 0$ signifies when an observed sequence is not encoded and used as it is.


We experiment with $r=0$ (no replication) and $r=1$ (one replication) to evaluate whether providing a pseudo-bidirectional context helps the LLM better identify causal patterns, especially in shorter sequence ($\leq 32$ symbols).

We first conduct an ablation study on synthetic data (Section~\ref{sec:results:synthetic}) to select the hyperparameters ($r, K$) of Algorithm~\ref{alg:causal_discovery} for application to the real dataset, whose results are presented in Section~\ref{sec:results:real}.
The universal symbol set $U$ is the set of "all English alphabets (upper and lower cased) and decimal numbers" i.e. $U$ = \{$a\dots z, A\dots Z, 0 \dots 9$\}.

\subsection{Data}
\label{sec:datasets}
To empirically validate our approach, we evaluated it on five datasets, described below. 
One  is generated synthetically (Section \ref{sec:syntheticdata}), and the remaining four consist of real-world time series. Since a time series is a continuous real valued sequence ordered temporally, it can also be viewed as a sequence of events by transforming temporal measurements into symbolic states, transitions, or occurrences. We use a simple transformation to extract symbolic event sequences from these time series data (Section \ref{subsec:real_data set}).

For both synthetic and real data, the symbols of the event sequences were arbitrarily chosen (Table \ref{tab:synthetic_data_stat} and \ref{tab:real_data_stat}) and hence do not hold any semantic meaning.

\subsubsection{Synthetic data}
\label{sec:syntheticdata} We use AGDES \cite{AGDES} to generate synthetic dependent pairs of encoded event sequences ($X, Y$) in single observation of varying lengths, where $(X, Y)$=$(X^{(0)}, Y^{(0)})$ (Sec. \ref{sec:methodology}) that are passed to Algorithm~\ref{alg:causal_discovery}. AGDES works by first generating a random Deterministic Finite Automaton (DFA) to produce the sequence $X$ of arbitrary length. This sequence is then processed sequentially by a randomly constructed Deterministic Finite Transducer (DFT) to generate the corresponding output sequence $Y$. Because the DFT's state transitions and outputs are strictly driven by the DFA's output sequence, this approach establishes a clear, one-way causal dependency from the input to the output. 
Table \ref{tab:synthetic_data_stat} summarizes the generated data which we used for evaluation. 
Even if some length ranges in Table \ref{tab:synthetic_data_stat} are not disjoint, each sequence pair for all subsets (Table \ref{tab:synthetic_data_stat}) is uniquely generated with distinct DFT and DFA, ensuring a strict single observation setting. For e.g. AGDES generates the 
pair 
(ab1dfha, lg0prq) of lengths $7$ and $6$ respectively.


\begin{table}[t]
\caption{
Number of synthetic sequence pairs $(X, Y)$ generated for each {causal direction} ($X \rightarrow Y$ or $Y \rightarrow X$) and {length range} (number of symbols per sequence).
Sequence symbols are chosen randomly from the set of all lower cased English alphabet and decimal numbers $(\{a\dots z, 0 \dots 9\})$.}
\label{tab:synthetic_data_stat}
\centering
\small
\setlength{\tabcolsep}{5pt}
\begin{tabular}{lcccc}
\hline
& \multicolumn{4}{c}{\textbf{Length range} (\# of symbols) $(|X|,|Y|)$} \\
\cmidrule(lr){2-5}
\textbf{Class} & 16--41 & 32--83 & 128--166 & 256--332 \\
\hline
$X \rightarrow Y$ & 376 & 156 & 101 & 99  \\
$Y \rightarrow X$ & 372 & 139 & 99  & 111 \\
\hline
\end{tabular}
\end{table}

\subsubsection{Real data}
\label{subsec:real_data set}
For testing on real event sequences, we rely on the publicly available IT-monitoring data introduced in \cite{aitbachir2023itmonitoring}\footnote{\url{https://github.com/ckassaad/Case_Studies_of_Causal_Discovery_from_IT_Monitoring_Time_Series}}, which 
consists of multivariate time series of monitoring metrics (e.g., CPU and RAM usage, disk reads/writes, network throughput etc.). 
The collection covers four subsets. 
1) \textit{MoM} (Middleware-oriented Message) contains 7 time series defined through the monitoring of an IT pipeline which ingests incoming messages based on a Publish/Subscribe architecture. 
2) \textit{Storm Ingestion} monitors a Storm ingestion system processing incoming messages. It consists of 8 time series detailing the inputs and outputs of different processes. 
3) \textit{Web Activity} consists of 10 time series 
reflecting the load on a web server. 
4) \textit{Antivirus Activity} consists of 13 time series 
reflecting the impact of antivirus activity in servers. 

For all the subsets we use their “version 1” from \cite{aitbachir2023itmonitoring}. 
Crucially for our setting, each subset comes with a \emph{ground truth summary causal graph}, 
from which we extract every \emph{direct} causal edge $X \rightarrow Y$ and treat the corresponding pair of time series $(X, Y)$ as one observed cause–effect pair. We exclude \emph{indirect} pairs (nodes connected only through intermediate variables) 
because they introduce hidden mediators and can carry causal signal, making direction recovery from single observations substantially harder.
Restricting to direct edges therefore provides the cleanest setting for evaluating our \emph{Sequence Generation Rule} (Section \ref{sec:llm_precedence_estimation}). 

\paragraph{Transformation to discrete event sequences:}

To convert the real valued IT monitoring time series to discrete event sequences we apply the following simple symbol-transition-based discretization technique. 

We conduct a 'preliminary processing' where we process only the non-stationary MoM subset using first-order differencing, moving-average smoothing ($w=3$), and standardization
. The stationary subsets (\textit{Storm Ingestion}, \textit{Web Activity}, and \textit{Antivirus Activity}) require no processing.

Then each series is independently partitioned into three discrete states using $q=3$ empirical quantiles. Rather than using the raw state sequences, we define an \emph{event} strictly as a \emph{state transition} (i.e., a change in the discretized quantile bin). This yields a maximum of $q(q-1) = 6$ distinct transition events per variable. These transition events are mapped to a symbol set. 

We use 
$q=3$ to control the alphabet size ($ |\Sigma| $, Section \ref{subsec:encoding}). Since some extracted event sequences are relatively short ($\approx 16$ symbols), $q=3$ ensures a sufficiently dense ``sequence-length to alphabet'' ratio to help the language model effectively recognize structural patterns via in-context learning. 

Ultimately, for each 
causal pair $(X, Y)$ of variables in the subsets, this produces the single observed encoded sequence pair $(X^{(0)}, Y^{(0)})$ (Sec. \ref{sec:methodology}) passed to Algorithm~\ref{alg:causal_discovery}.

Table \ref{tab:real_data_stat} describes the discretized real data used. For the \textit{Antivirus Activity} and \textit{Web Activity} datasets, the sequence obtained were too long, so we use a shorter sequence (of length $\approx 1/3$ of the total). 

\textcolor{red}{To remove the one-label ($X \to Y$) bias in Table \ref{tab:real_data_stat}, each pair evaluation (Sec. \ref{sec:results:real}) 
uses 10 balanced pair-and-label direction randomizations 
where five retain ($X,Y$) with $X \to Y$ label, and five reverse to ($Y,X$) with $Y \to X$.
}

\begin{table}[t]
\caption{Statistics of discretized real-world datasets: number of sequence pairs $(X, Y)$ (all with $X \rightarrow Y$ causal direction) for each subset, including length ranges $(|X|, |Y|)$.
The event sequence symbols are randomly chosen from the set \{a,b,c,d,e,f,A,B,C,D,E,F\}.}
\label{tab:real_data_stat}
\centering
\small
\setlength{\tabcolsep}{4pt}
\begin{tabular}{lccc}
\hline
\textbf{Subset} &
\textbf{\begin{tabular}[c]{@{}c@{}}Length Range\\ $(|X|,|Y|)$\end{tabular}} &
\textbf{Class} &
\textbf{\# Pairs} \\
\hline
\textit{MoM}                & 14--191 & $X \rightarrow Y$ & 10 \\
\textit{Storm Ingestion}    & 52--170 & $X \rightarrow Y$ & 9  \\
\textit{Web Activity}       & 46--1000 & $X \rightarrow Y$ & 14 \\
\textit{Antivirus Activity} & 19--256 & $X \rightarrow Y$ & 16 \\
\hline
\end{tabular}
\end{table}

\subsubsection{Pre-processing event sequences}
\label{subsubsec:pre-processing_event_sequences}
Since modern LLMs use subword tokenizers (such as Byte-Pair Encoding) that may merge contiguous characters into single tokens, we format the event sequences 
prior to inference, where we apply ``uniform spacing'' by inserting whitespace between symbols (e.g., `A B C' instead of `ABC'). 
This is done to properly leverage pre-trained LLMs for our causal scoring framework. 

Spacing ensures each event is treated as an independent token, guaranteeing that the causal scores are computed at the precise symbol/token granularity of the individual sequence events.

\subsection{Baseline methods}
\label{sec:baselines}





We compare our framework against two families of baselines: Standard LLM prompting for both synthetic and real data, and classical causal inference algorithms for real data only.

\subsubsection{Zero-shot LLM prompting baselines (synthetic \& real data)}
\label{subsec:zeroshot_baseline}
To isolate the contribution of our method (using the \emph{Sequence Generation Rule}), we use standard zero-shot prompting 
and directly query the same backbone LLMs (Section~\ref{sec:experimental_setup:model}). In addition we also use bigger
LLMs (Llama-3.1-8B-Instruct, Llama-3.3-70B-Instruct) which are more capable than the ones used in the backbone of our Algorithm \ref{alg:causal_discovery}) for standard zero-shot prompting baseline. 
To the best of our knowledge, since classical causal inference methods are not applicable in the setting of single observed discrete event sequences, this is the only applicable existing baselines for this setting.

As per the data (Section \ref{sec:datasets}) and our problem definition (Section \ref{sec:methodology}), the encoding of the event sequences do not hold any semantic meaning, hence to have a fair comparison with our method we experiment with three variations of the standard zero-shot prompting baseline: (a) (No encoding, $r=0$): the standard prompting where the sequences are used as it is without replication in the prompt; (b) (Encoded and $r=0$): to show the effect of random sequence encoding; 
(c) (Encoded and $r=1$): to show the effect of random sequence encoding and replication in the prompt. 

Separately to emphasize the advantage of our method over the standard zero-shot prompting we also experimented with max-voting over $K \in \{51, 311\}$ encodings for the (b) (Encoded and $r=0$) and (c) (Encoded and $r=1$) variations of zero-shot prompting above. 

\begin{table}[t]
\centering
\caption{Average test performance ($\mu$-F1) on Synthetic data  across all sequence lengths. $\dagger$ indicates the Instruct version of the model. 
Results compare three settings: (a) classical zero-shot LLM prompting,
    (b) zero-shot with one random encoding (\textit{Encoded}), and
    (c) zero-shot with one random encoding and one replication ($r=1$); with our method ($K=1111, r=1$). }
\label{tab:main_comparison}
\setlength{\tabcolsep}{6pt} 
\begin{tabular}{llc}
\toprule
\multicolumn{2}{l}{\textbf{Model / Setting}} & \textbf{$\mu$-F1} \\
\midrule
\multirow{11}{*}{\rotatebox[origin=c]{90}{\text{Standard prompting baseline}}} 
& \multicolumn{2}{l}{\textit{(a) Zero-Shot (No encoding, $r=0$)}} \\
& Llama-3.2-3B$^\dagger$ & 0.4868 \\
& Llama-3.1-8B$^\dagger$ & 0.5016 \\
& Llama-3.3-70B$^\dagger$ & 0.5110 \\
\cmidrule{2-3}
& \multicolumn{2}{l}{\textit{(b) Zero-Shot (Encoded, $r=0$)}} \\
& Llama-3.2-3B$^\dagger$ & 0.4901 \\
& Llama-3.1-8B$^\dagger$ & 0.5057 \\
& Llama-3.3-70B$^\dagger$ & 0.4972 \\
\cmidrule{2-3}
& \multicolumn{2}{l}{\textit{(c) Zero-Shot (Encoded, $r=1$)}} \\
& Llama-3.2-3B$^\dagger$ & 0.4930 \\
& Llama-3.1-8B$^\dagger$ & 0.5234 \\
& Llama-3.3-70B$^\dagger$ & 0.4984 \\
\midrule
\multicolumn{2}{l}{\textbf{Our Method ($K=1111, r=1$)}} & \\
\multicolumn{2}{l}{\textbf{Llama-3.2-3B}} & \textbf{0.9194} \\
\bottomrule
\end{tabular}
\end{table}

\subsubsection{Classical causal discovery baselines (real data only)}
\label{subsubsec:classical_baselines}

Because the IT-monitoring datasets are originally real-valued time series, standard time-series causal discovery algorithms can also be applied to them. On the (\emph{undiscretized}) time series of each direct causal pair, we additionally compare against three popular methods: Granger Causality  \cite{granger_causality}, \textit{VARLiNGAM} \cite{hyvarinen2010lingam} and \textit{TiMINo} \cite{peters2013timino}. 
%
They are run with the hyper-parameters reported in \cite{aitbachir2023itmonitoring}: linear kernels / models, significance threshold = $0.05$, and a maximum lag of 10 time steps.

Except Granger Causality, these are \emph{multi-observation} algorithms by design, and all use the entire numerical time series of each pair. 
So they receive strictly more information than our method, which uses only one discretized event sequence pair. Any subsequent comparison is therefore conservative with respect to our approach.

Unlike our binary classification method, classical baselines may output 'independence' or ``undecided'' (bidirectional or unoriented). 
For fair comparison, we randomly assign a direction to non-oriented outputs (with $p=0.5$) and average the results over 10 trials to reduce variance (Table~\ref{tab:real_data_results}).


\subsubsection{Evaluation metrics:}
Since our problem is a binary classification problem, with the two classes being the two possible orientations, we report micro-averaged F1 ($\mu$-F1) score. In the binary setting, Accuracy, $\mu$-Precision, $\mu$-Recall, and $\mu$-F1 are all equal.

\section{Experimental results}
\label{sec:results}
We analyze the performance of our method and baselines over synthetic data 
and 
real data in Section \ref{sec:results:synthetic} and \ref{sec:results:real}
below.

\subsection{Results on synthetic data}
\label{sec:results:synthetic}
\subsubsection{General conclusion} 
Table \ref{tab:main_comparison} compares the average performance of our method (using Llama-3.2-3B as the backbone, with $K=1111$ and $r=1$) against the standard Zero-Shot baselines across three different instruction-tuned models, averaged over all four length ranges. 
While standard zero-shot prompting yields near-random performance ($50\%$ $\mu$-F1), even for large models like Llama-3.3-70B, our Sequence Generation Rule-based algorithm (Algorithm \ref{alg:causal_discovery}) achieves an average $\mu$-F1 of $91.94\%$
Although replication in `Encoded' settings provides minor improvements for baselines, all zero-shot prompting baselines remain near random. 
In contrast, our approach outperforms all zero-shot prompting baselines by a wide margin 
for inferring causal direction in single observation.

\begin{table}[t!]
\centering
\caption{Effect of sequence lengths and replication ($r \in \{0,1\}$) on performance of our method (with Llama-3.2-3B, $K=1111$) on synthetic data. 
}
\label{tab:length_effect}
\begin{tabular}{ccc}
\hline
\multirow{2}{*}{\textbf{\begin{tabular}[c]{@{}c@{}}Length Range\\ $(|X|,|Y|)$\end{tabular}}} & \multicolumn{2}{c}{\textbf{$\mu$-F1}} \\ \cline{2-3} 
                                                                                             & ($r=0$)                 & ($r=1$)                \\ \hline
16-41                                                                                        & 0.5856                  & 0.7888                 \\
32-83                                                                                        & 0.5153                  & 0.9322                 \\
128-166                                                                                      & 0.9450                  & 0.9900                 \\
256-332                                                                                      & 0.9571                  & 0.9667                 \\ \hline
Avg.                                                                                         & 0.7477                  & 0.9194                
\end{tabular}
\end{table}

Additionally, a lateral set of experiments (Sec.~\ref{subsec:zeroshot_baseline}) not detailed here in Table \ref{tab:main_comparison} used max-voting over ($K \in \{51,311\}$) encodings for the zero-shot prompting baselines: (b) Encoded with (r=0) and (c) Encoded with (r=1). These variants produced similar across all tested LLMs, with near-random $\mu$-F1 scores ranging between $0.4-0.6$, comparable to the baseline in Table \ref{tab:main_comparison}.

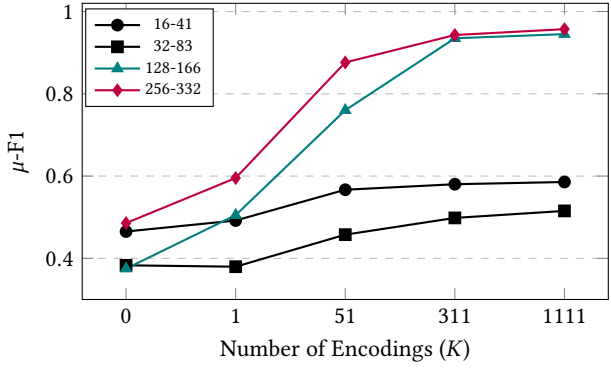
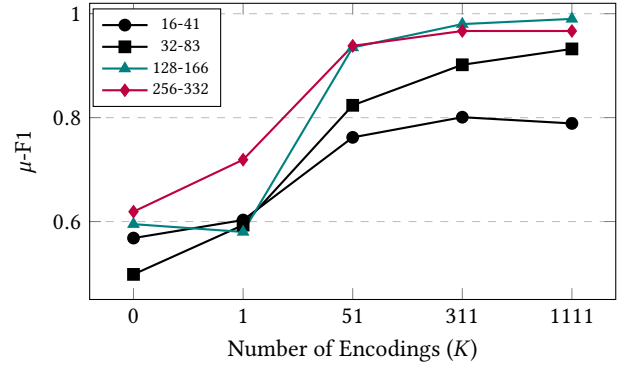
\begin{figure*}[t]
    \centering
    \begin{subfigure}{0.48\textwidth}
        \centering
        \begin{tikzpicture}
            \begin{axis}[
                title={   },
                xlabel={Number of Encodings ($K$)},
                ylabel={$\mu$-F1},
                symbolic x coords={0, 1, 51, 311, 1111},
                xtick={0, 1, 51, 311, 1111},
                ymin=0.3, ymax=1.02,
                legend style={at={(0.25,0.65)}, anchor=south east, nodes={scale=0.75, transform shape}},
                ymajorgrids=true,
                grid style=dashed,
                width=\textwidth,
                height=5.5cm
            ]
            
            \addplot[color=blue, mark=*, thick] coordinates {
                (0, 0.4652)
                (1, 0.4920)
                (51, 0.5668)
                (311, 0.5802)
                (1111, 0.5856)
            };
            \addlegendentry{16-41}
            
            \addplot[color=red, mark=square*, thick] coordinates {
                (0, 0.3831)
                (1, 0.3797)
                (51, 0.4576)
                (311, 0.4983) 
                (1111, 0.5153)
            };
            \addlegendentry{32-83}
            
            \addplot[color=teal, mark=triangle*, thick] coordinates {
                (0, 0.3750)
                (1, 0.5050)
                (51, 0.7600)
                (311, 0.9350) 
                (1111, 0.9450)
            };
            \addlegendentry{128-166}
            
            \addplot[color=purple, mark=diamond*, thick] coordinates {
                (0, 0.4857)
                (1, 0.5952)
                (51, 0.8762)
                (311, 0.9429)
                (1111, 0.9571)
            };
            \addlegendentry{256-332}
            
            \end{axis}
        \end{tikzpicture}
        \caption{Performance without replication ($r=0$).}
        \label{subfig:k_mappings_no_rep}
    \end{subfigure}\hfill
    \vspace{6pt}
    \begin{subfigure}{0.48\textwidth}
        \centering
        \begin{tikzpicture}
            \begin{axis}[
                title={   },
                xlabel={Number of Encodings ($K$)},
                ylabel={$\mu$-F1},
                symbolic x coords={0, 1, 51, 311, 1111},
                xtick={0, 1, 51, 311, 1111},
                ymin=0.45, ymax=1.02,
                legend style={at={(0.25,0.65)}, anchor=south east, nodes={scale=0.75, transform shape}},
                ymajorgrids=true,
                grid style=dashed,
                width=\textwidth,
                height=5.5cm
            ]
            
            \addplot[color=blue, mark=*, thick] coordinates {
                (0, 0.5682)
                (1, 0.6029)
                (51, 0.7620)
                (311, 0.8008)
                (1111, 0.7888)
            };
            \addlegendentry{16-41}
            
            \addplot[color=red, mark=square*, thick] coordinates {
                (0, 0.4983)
                (1, 0.5932)
                (51, 0.8237)
                (311, 0.9017) 
                (1111, 0.9322)
            };
            \addlegendentry{32-83}
            
            \addplot[color=teal, mark=triangle*, thick] coordinates {
                (0, 0.5950)
                (1, 0.5800)
                (51, 0.9350)
                (311, 0.9800) 
                (1111, 0.9900)
            };
            \addlegendentry{128-166}
            
            \addplot[color=purple, mark=diamond*, thick] coordinates {
                (0, 0.6190)
                (1, 0.7190)
                (51, 0.9381)
                (311, 0.9667)
                (1111, 0.9667)
            };
            \addlegendentry{256-332}
            
            \end{axis}
        \end{tikzpicture}
        \caption{Performance with replication ($r=1$).}
        \label{subfig:k_mappings_with_rep}
    \end{subfigure}
    \caption{Curves showing performance variation of our method (with Llama-3.2-3B) as $K$ varies between 0 (raw observed sequence) and 1111, for different sequence lengths.}
    \label{fig:k_mappings_detailed}
\end{figure*}

\subsubsection{Effect of sequence lengths}
\label{subsubsec:effect_sequence_length_synthetic}
To understand how sequence lengths affect Algorithm \ref{alg:causal_discovery}, we evaluate the performance of our method across different sequence length ranges in Table \ref{tab:length_effect}. 
%
%
This table clearly shows that sequence length impacts the predictive performance of our Algorithm \ref{alg:causal_discovery}, regardless of replication.\\
 \textit{Without replication ($r=0$):} Inferring causal direction purely from raw event sequences struggles significantly with short contexts. For sequences of length ($|X|$ or $|Y|$) between $16-41$ \& $32-83$ symbols, the performance ($\mu$-F1) hovers near random guessing ($0.5153$ to $0.5856$). However, with longer sequence length ($128-332$ symbols), the LM successfully recognizes the underlying latent patterns, jumping over $0.94$ $\mu$-F1 score.\\
 \textit{With replication ($r=1$):} Applying just one replication substantially mitigates the short-sequence penalty. For sequence lengths $32-83$, $\mu$-F1 jumps from $0.5153$ to $0.9322$. However, the influence of sequence length persists even when $r=1$; the performance starts at $0.7888$ $\mu$-F1 for the shortest sequences ($16-41$), jumps to $0.9900$ $\mu$-F1 for sequences of length $128-166$, and $0.9667$ for sequences of length $256-332$. 
    

\subsubsection{Effect of replications}
\label{subsubsec:effect_replication_synthetic}
As seen from Table \ref{tab:length_effect} above, increasing context through replication ($r=1$) for our Algorithm \ref{alg:causal_discovery} 
significantly aids the language model in better sequence prediction which ultimately helps to get better $S$ scores. When averaged across all lengths at $K=1111$, Llama-3.2-3B achieves a $\mu$-F1 of 0.9194 with replication ($r=1$), compared to 0.7477 without replication ($r=0$). This improvement is most evident in the shorter ($<83$ length) subsets. This shows that replication helps the LLM identify patterns for better sequence modelling, specially for short sequences.

\subsubsection{Effect of encodings}
\label{subsubsec:effect_encoding_synthetic}

To isolate the impact of our randomized causal encoding strategy over unencoded sequences, we assess how the performance scales as we increase the number of encodings ($K$) in our method (with Llama-3.2-3B). Figure \ref{fig:k_mappings_detailed} visualizes this impact against sequence length ranges for both the standard ($r=0$) and the replicated ($r=1$) settings. Here,
 $K=0$ represents the raw sequence text (unencoded baseline),
 $K=1$ represents a single random encoding (no averaging over scores),
 $K > 1$ represents our proposed method of taking expectation (Eq. \ref{eq:expectation_over_scores}) over scores across multiple random encoding mappings.


Below, we analyze the transition trend in Figure \ref{fig:k_mappings_detailed}, by comparing our method with $K>1$ against $K=0$ and $K=1$, highlighting the importance of the encoding process.
 
\paragraph{Challenges with raw event sequences ($K=0$):} Algorithm \ref{alg:causal_discovery} struggle significantly with raw event sequences. Without replication ($r=0$, $K=0$), performance falls below $0.5$ $\mu-$F1 score across all lengths, occasionally dropping as low as $0.375$ $\mu$-F1 (for length $128-166$). This poor performance is likely caused by semantic biases inherent to the LLM, which are unintentionally inherited through the choice of the event symbols and their arrangements during sequence observation and generation. Even with replication ($r=1$), the $\mu$-F1 score is bounded between $0.49$ and $0.62$.
    
\paragraph{The volatility of single encodings ($K=1$):} Introducing a single random encoding ($K=1$) remaps tokens into an alternative symbolic alphabet. The resulting variability between $K=0$ and $K=1$ in our method's performance (Figure \ref{fig:k_mappings_detailed}) indicates that LLM-based inference is highly sensitive to the specific choice of sequence symbols. 
This suggests that for our problem approaches utilizing LLMs (such as zero-shot prompting baselines) and relying on only a few randomly sampled encoding can be unstable, with performance gains that are neither consistent nor robust across settings. 
For example, in Figure \ref{subfig:k_mappings_no_rep} (under $r=0$) for length $128$–$166$, performance increases from $0.3750$ to $0.5050$ when moving from $K=0$ to $K=1$. In contrast, under $r=1$ for the same length range, performance decreases from $0.5950$ to $0.5800$. 

Similar performance variations for our method on synthetic data, for $K \in \{0,1\}$, can also be seen with other LLMs in Table~\ref{tab:lm_choice_raw}.

\paragraph{Performance scaling and convergence ($K \ge 51$):} A significant performance gap is covered after $K \ge 51$, where in both plots in Figure \ref{fig:k_mappings_detailed}, the performance increases steeply from the unstable $K=0$ and $K=1$ region. For lengths $\ge 128$ under $r=0$, the $\mu$-F1 score climbs smoothly until it plateaus. Under $r=1$, this ascent is even steeper where the performance with sequences of $128-166$ length, for example, improves from $0.5800$ $\mu-$F1 ($K=1$) to $0.9350$ $\mu-$F1 ($K=51$), ultimately stabilizing at $0.9900$ $\mu-$F1 ($K=1111$). For most length ranges, the greatest marginal gain is achieved by expanding the mapping pool to around $K=311$. Moving from $K=311$ to $K=1111$ yields smaller improvements but are more statistically stable as the expectation is over a larger pool. Hence for testing on the real data we use a higher $K=1111$. Similar performance variations with $K\geq 51$, can also be seen with other LLMs, in Table~\ref{tab:lm_choice_raw}.
    
\paragraph{Mitigating LLM token-bias:} Nearly across all settings, scaling the number of token encodings from $K=51$ to $K=311$ yields a sharp increase in performance. Language models have strong pre-trained semantic biases for specific tokens (e.g., specific English words or numbers). 
This demonstrates how expectation of causal scores $S$ (Eq. \ref{eq:expectation_over_scores}) over hundreds of distinct randomized encodings effectively mitigates token-specific modeling bias, leaving only the true structural and syntactic sequence signals.
Similar evidences can also be seen with other LLMs, in Table~\ref{tab:lm_choice_raw}, for $K \geq 51$, where the performance of our method becomes more stable at higher $K$.
     
\paragraph{Bottleneck in non-replicated ($r=0$) sequences:} In Figure \ref{subfig:k_mappings_no_rep} (under $r=0$), expanding $K$ drastically improves the $\mu-$F1 performance for long sequences, jumping from $0.76$ (at $K=51$) to $0.945$ (at $K=1111$) for lengths $128-166$, and from $0.8762$ (at $K=51$) to $0.9571$ (at $K=1111$) for lengths $256-332$. However, the performance with shorter sequences (lengths $16-41$ and $32-83$) remain largely stagnant near the $0.50-0.58$ $\mu-$F1 mark, regardless of how many encodings ($K$) are evaluated. This indicates that for our method at $r=0$, the performance limiting factor for these short sequences is not LLM token bias, but the lack of enough structural information for in-context learning in the unidirectional causal attention. 

\paragraph{Combination of $K$ and replication ($r=1$):} Figure \ref{subfig:k_mappings_with_rep} shows the behaviour of replication in combination with $K$ in our method. By replicating the sequence just once, the method receives enough context to remove the ``short sequence'' bottleneck by a huge margin. Once the sequences are replicated, increasing the number of encoding $K$ yields further performance improvements. This is most notably observed in the $32-83$ length range, where increasing $K$ from $51$ to $1111$ improves the performance from $0.8237$ up to $0.9322$ $\mu-$F1 score. A similar trend is also evident with other backbone LLM for our method (Table~\ref{tab:lm_choice_raw}) for $K \geq 51$ and $r=1$.

\begin{table}
\centering
\caption{Detailed performance ($\mu$-F1 scores) 
on synthetic data using GPT-2 and Llama-3.2-1B backbones. 
$\dagger$ = sequence truncated to 256 symbol length for GPT-2 context limit.}
\label{tab:lm_choice_raw}
\small
\setlength{\tabcolsep}{3.2pt} 
\begin{tabular}{llc ccccc}
\toprule
\multirow{2}{*}{\textbf{\begin{tabular}[c]{@{}l@{}}Our \\ Method\end{tabular}}} & \multirow{2}{*}{\textbf{$r$}} & \multirow{2}{*}{\textbf{Length}} & \multicolumn{5}{c}{\textbf{\# of Encodings ($K$)}} \\
\cmidrule(lr){4-8}
& & & \textbf{0} & \textbf{1} & \textbf{51} & \textbf{311} & \textbf{1111} \\ 
\midrule
\textbf{with} & $r=0$ & 16-41 & 0.5548 & 0.5495 & 0.5468 & 0.5508 & 0.5388 \\
\textbf{GPT-2}& & 32-83 & 0.5186 & 0.5458 & 0.5017 & 0.4678 & 0.4475 \\
& & 128-166 & 0.4550 & 0.6450 & 0.7100 & 0.7000 & 0.7200 \\
& & 256-332 & 0.5238 & 0.6619 & 0.6286 & 0.6952 & 0.6619 \\ 
\cmidrule(lr){3-8}
& & \textit{Avg.} & \textit{0.5131} & \textit{0.6005} & \textit{0.5968} & \textit{0.6035} & \textit{0.5920} \\ 
\cmidrule(lr){2-8}
& $r=1$ & 16-41 & 0.5936 & 0.5602 & 0.6203 & 0.6230 & 0.6872 \\
& & 32-83 & 0.4814 & 0.4949 & 0.3966 & 0.3932 & 0.7390 \\
& & 128-166 & 0.5800 & 0.5350 & 0.7500 & 0.7500 & 0.7800 \\
& & 256-332 & 0.6095$^\dagger$ & 0.7046$^\dagger$ & 0.8476$^\dagger$ & 0.8857$^\dagger$ & 0.8762$^\dagger$ \\ 
\cmidrule(lr){3-8}
& & \textit{Avg.} & \textit{0.5516} & \textit{0.5736} & \textit{0.6536} & \textit{0.6629} & \textit{0.7706} \\ 
\midrule
\textbf{with} & $r=0$ & 16-41 & 0.4612 & 0.5027 & 0.6872 & 0.6270 & 0.6243 \\
\textbf{Llama} & & 32-83 & 0.3051 & 0.5932 & 0.7458 & 0.6441 & 0.6542 \\
\textbf{3.2 (1B)}& & 128-166 & 0.5700 & 0.5800 & 0.9250 & 0.8900 & 0.9500 \\
& & 256-332 & 0.6857 & 0.6333 & 0.9429 & 0.9714 & 0.9762 \\ 
\cmidrule(lr){3-8}
& & \textit{Avg.} & \textit{0.5055} & \textit{0.5773} & \textit{0.8252} & \textit{0.7831} & \textit{0.8012} \\ 
\cmidrule(lr){2-8}
& $r=1$ & 16-41 & 0.7754 & 0.6270 & 0.7754 & 0.7300 & 0.7513 \\
& & 32-83 & 0.5966 & 0.7119 & 0.9458 & 0.8949 & 0.9051 \\
& & 128-166 & 0.7550 & 0.7000 & 0.9450 & 0.9950 & 0.9850 \\
& & 256-332 & 0.6810 & 0.6476 & 0.9762 & 0.9800 & 0.9850 \\ 
\cmidrule(lr){3-8}
& & \textit{Avg.} & \textit{0.7020} & \textit{0.6716} & \textit{0.9106} & \textit{0.8999} & \textit{0.9066} \\ 
\bottomrule
\end{tabular}
\end{table}

\begin{table*}[htbp]
\centering
\caption{\textcolor{red}{Performance ($\mu$-F1, mean $\pm$ standard deviation) of Algorithm~\ref{alg:causal_discovery} on real data, after 10 balanced pair-and-label randomization. 
Zero-shot baseline uses discretized sequences with random encoding for $r\in${0,1} across 50 iterations. Our method (with K=1111) is run over four random seeds.
$\dagger$ indicates instruct version of the respective LLM}}
\label{tab:real_data_results}
\footnotesize
\setlength{\tabcolsep}{4pt}
\begin{tabular}{llcccc}
\hline
\textbf{Category} & \textbf{Method / Model} & \textit{\textbf{Storm Ingestion}} & \textit{\textbf{Antivirus Activity}} & \textit{\textbf{MoM}} & \textit{\textbf{Web Activity}} \\
\hline
\multirow{3}{*}{\textit{Classical baselines}} & Granger Causality & \textcolor{red}{$0.558_{\pm{0.167}}$} & \textcolor{red}{$0.494_{\pm{0.083}}$} & \textcolor{red}{$0.342_{\pm{0.135}}$} & \textcolor{red}{$0.555_{\pm{0.113}}$} \\
& VARLiNGAM & \textcolor{red}{$0.450_{\pm{0.102}}$} & \textcolor{red}{$0.572_{\pm{0.106}}$} & \textcolor{red}{$0.439_{\pm{0.105}}$} & \textcolor{red}{$0.357_{\pm{0.087}}$} \\
& TiMINo & \textcolor{red}{$0.552_{\pm{0.134}}$} & \textcolor{red}{$0.565_{\pm{0.126}}$} & \textcolor{red}{$0.391_{\pm{0.133}}$} & \textcolor{red}{$0.501_{\pm{0.118}}$} \\
\hline
\multirow{3}{*}{\begin{tabular}[c]{@{}l@{}}\textit{\textcolor{red}{Zero-Shot prompting baseline}}\\\textit{\textcolor{red}{(Encoded, $r=0$; 50 iterations
)}}\end{tabular}}
& \textcolor{red}{Llama-3.2-3B$^\dagger$} & \textcolor{red}{$0.502_{\pm{0.163}}$} & \textcolor{red}{$0.499_{\pm{0.126}}$} & \textcolor{red}{$0.498_{\pm{0.161}}$} & \textcolor{red}{$0.504_{\pm{0.124}}$} \\
& \textcolor{red}{Llama-3.1-8B$^\dagger$} & \textcolor{red}{$0.502_{\pm{0.161}}$} & \textcolor{red}{$0.509_{\pm{0.121}}$} & \textcolor{red}{$0.502_{\pm{0.155}}$} & \textcolor{red}{$0.500_{\pm{0.138}}$} \\
& \textcolor{red}{Llama-3.3-70B$^\dagger$} & \textcolor{red}{$0.500_{\pm{0.162}}$} & \textcolor{red}{$0.495_{\pm{0.122}}$} & \textcolor{red}{$0.500_{\pm{0.159}}$} & \textcolor{red}{$0.503_{\pm{0.139}}$} \\
\hline
\multirow{3}{*}{\begin{tabular}[c]{@{}l@{}}\textit{\textcolor{red}{Zero-Shot prompting baseline}}\\\textit{\textcolor{red}{(Encoded, $r=1$; 50 iterations
)}}\end{tabular}}
& \textcolor{red}{Llama-3.2-3B$^\dagger$} & \textcolor{red}{$0.499_{\pm{0.157}}$} & \textcolor{red}{$0.498_{\pm{0.123}}$} & \textcolor{red}{$0.493_{\pm{0.157}}$} & \textcolor{red}{$0.489_{\pm{0.117}}$} \\
& \textcolor{red}{Llama-3.1-8B$^\dagger$} & \textcolor{red}{$0.501_{\pm{0.171}}$} & \textcolor{red}{$0.494_{\pm{0.121}}$} & \textcolor{red}{$0.487_{\pm{0.161}}$} & \textcolor{red}{$0.501_{\pm{0.137}}$} \\
& \textcolor{red}{Llama-3.3-70B$^\dagger$} & \textcolor{red}{$0.506_{\pm{0.163}}$} & \textcolor{red}{$0.493_{\pm{0.113}}$} & \textcolor{red}{$0.485_{\pm{0.162}}$} & \textcolor{red}{$0.524_{\pm{0.133}}$} \\
\hline
\textbf{Ours (Llama-3.2-3B)} & ($r = 1, K=1111$) & \textcolor{red}{\textbf{$0.611_{\pm{0.056}}$}} & 
\textcolor{red}{\textbf{$0.781_{\pm{0.031}}$}} & 
\textcolor{red}{\textbf{$0.725_{\pm{0.043}}$}} & 
\textcolor{red}{\textbf{$0.696_{\pm{0.031}}$}} 
\\
\hline
\end{tabular}
\end{table*}

\subsubsection{Impact of the backbone LLM}

We extend the evaluation of our Algorithm \ref{alg:causal_discovery} using two other backbone LLMs, GPT-2 and Llama-3.2-1B  in detail in Table \ref{tab:lm_choice_raw}. 
It reveals that our causal direction discovery framework depends on the underlying LLM capability, with a clear gap between older models like GPT-2 and modern Llama-3.2 variants. Even under optimal settings ($r=1, K=1111$), GPT-2 peaks at a mean $\mu$-F1 of $0.7706$, while Llama-3.2 models consistently exceed a $\mu$-F1 score of $0.90$. This difference indicates that our method fundamentally relies on strong in-context learning and well-developed induction heads in modern LLMs for identifying structural patterns without fine-tuning for our out-of-distribution event sequences.

Despite the raw performance differences between GPT-2 and Llama models, the performance trends across sequence lengths (Section \ref{subsubsec:effect_sequence_length_synthetic}), replication $r$ (Section \ref{subsubsec:effect_replication_synthetic}) and encodings $K$ (Section \ref{subsubsec:effect_encoding_synthetic}) remain consistent. Across all evaluations, sequence replication ($r=1$) and a large ($K$) universally improves performance. 

Llama-3.2-1B performs strongly, with an average $\mu$-F1 of $0.9066$ ($r=1, K=1111$), close to Llama-3.2-3B’s $\mu$-F1 of $0.9194$ (Table \ref{tab:length_effect}).  Notably, the 1B model slightly exceeds the 3B model in the $r=0$ setting. However, we use Llama-3.2-3B as the primary backbone due to its greater capacity yielding better in-context learning.

\subsection{Test results on real data}
\label{sec:results:real}

To evaluate generalization, we test our framework on real-world IT-monitoring datasets introduced in Section \ref{subsec:real_data set} as a strictly unseen test set, without any hyperparameter tuning. We directly apply the optimal configuration (\textit{Llama-3.2-3B}, $K=1111$, and $r=1$) obtained after evaluation on synthetic data. \textcolor{red}{Evaluations randomize the pairs and their ground-truth directions for removing the single label ($X\to Y$) bias (see Sec.~\ref{subsec:real_data set}).} 


\textcolor{red}{
The results demonstrate that our proposed method (with Llama-3.2-3B, $r=1, K=1111$), substantially outperforms both the classical causal-discovery methods and the zero-shot prompting baselines on most datasets. In particular, it achieves the highest $\mu$-F1 scores on Antivirus Activity (0.813), MoM (0.700), and Web Activity (0.714). The improvements for Antivirus Activity and MoM, are the strongest where the classical baselines reach only 0.572 and 0.439, respectively. On Storm Ingestion, our method obtains 0.556, which is competitive with the best classical result of 0.558 from Granger Causality and slightly exceeds TiMINo at 0.552. 
}

\textcolor{red}{
In contrast, the zero-shot prompting baselines remain close to 0.5 $\mu$-F1 across datasets and model sizes, with no evidence that increasing the Llama model size from 3B to 70B improves the performance on our pairwise causal direction discovery problem. This suggests that model scale alone is insufficient for our problem under the zero-shot prompting, whereas our proposed method enables a smaller Llama-3.2-3B model to extract more useful information for causal direction orientation. Moreover, the reported standard deviation of 0.000 for our method indicates no observed variation across the evaluations summarized in the table, while both the classical and zero-shot baselines exhibit high variability. \footnote{This variability in zero-shot baselines arise through the randomization of the encoding and prompt (through the causal pair swapping for removing the single $X\to Y$ label bias). For the classical baselines, it arises from the random binning of the non-oriented pairs (See Section \ref{subsubsec:classical_baselines}). For our method the only source of variation is through K which is chosen randomly, but averaging over all K encoding helps to balance the variations in favour of a more robust prediction.} In addition, a lateral set of experiments (Sec.~\ref{subsec:zeroshot_baseline}), not detailed here in Table \ref{tab:real_data_results}, used max-voting over ($K \in \{51,311\}$) encodings for the zero-shot prompting baselines: (b) Encoded with (r=0) and (c) Encoded with (r=1), gave similar near-random $\mu$-F1 performance ranging between $0.45-0.5$ (with variance $\pm 0.1-0.15$) for all LLM in Table \ref{tab:real_data_results}. Overall, these results suggest that algorithm \ref{alg:causal_discovery} offers both stronger predictive performance and greater empirical stability than the competing approaches for our causal direction discovery problem (Sec. \ref{sec:methodology}).
}


\section{Conclusion}
\label{sec:conclusion}

In this work, we study the problem of inferring the causal direction between two event sequences from a single observation, and propose a framework that leverages the sequence modeling and in-context learning capabilities of pre-trained LLMs as training-free probability density estimators. The framework is based on our \textit{Sequence Generation Rule}, which compares the negative log-likelihoods of causally oriented event sequences in both directions. To improve robustness in practice, we employ random bijective encodings to reduce semantic bias and sequence replication to strengthen in-context learning.

Our method achieves an average $\mu$-F1 of ~92\% on synthetic data, demonstrating robustness across various sequence lengths. On real-world IT-monitoring data, the framework identifies correct causal directions and significantly outperforms traditional baselines, such as Granger Causality, VARLiNGAM and TiMINo, despite these baselines having access to the full continuous time series. Our ablation studies further highlight the importance of both random encodings and replication 
for performance stability. 


\textcolor{blue}{These findings suggest that pre-trained sequence models can provide useful structural signals for orienting causal relations under single observation. This setting can arise in event and relationship mining over web-service logs, telecommunications, monitoring data, or other event streams where a candidate relation between two processes is known but its direction remains unresolved.}
Our future work will focus on incorporating temporal information into the scoring mechanism and extending this pairwise framework to recover full causal graphs.

\section{Limitations}
\label{sec:limitations}

\textcolor{blue}{The \emph{Sequence Generation Rule} and \emph{Rule A} determines causal direction under the assumption that a direct causal relation already exists between the two processes in the observed sequence pair. Therefore, Algorithm 1 always chooses one of the two possible directions and does not distinguish direct causation from independence, confounding, or indirect association. Identifying whether a direct causal link exists is a separate task and must be provided by prior knowledge or another inference step.
}

Algorithm \ref{alg:causal_discovery} depends on the raw probability estimations and hence we use the base versions of the LLMs (Llama-3.2-3B and Llama-3.2-1B), rather than their Instruct variants (Llama-3.2-3B-Instruct and Llama-3.2-1B-Instruct)  
which are specifically trained to be used with prompts/instructions.

The maximum sequence length that can be processed by our Algorithm \ref{alg:causal_discovery} is bounded by the maximum context limit of the LLM used. For a given pair of sequences with lengths $n$ and $m$, and a replication factor $r$ the input length ($\mathcal{L}$), becomes $(r+1)(n+m) \times 2$ (accounting for added space tokens, Section \ref{subsubsec:pre-processing_event_sequences}). $\mathcal{L} \leq $  max context limit of the LLM used (e.g., $1024$ for GPT-2).  

The computational complexity of Algorithm \ref{alg:causal_discovery} is dominated by the LLM.
Given the standard Transformer's $\mathcal{O}(L^2)$ self-attention mechanism, evaluating both causal directions for a single encoding takes $\mathcal{O}(L^2 \cdot d)$ time, where $d$ is the model's hidden dimension and $L$ is its input sequence length. Consequently, for our method, aggregating the causal scores ($S$) over $K$ encodings yields an overall time complexity of $\mathcal{O}\left(K \cdot \mathcal{L}^2 \cdot d\right)$ per pair. To maintain tractable inference times with this quadratic complexity we used a moderately sized LLM (e.g., Llama-3.2-3B) and truncated exceedingly long event sequences 
(Section \ref{subsec:real_data set} on real data).


Although our method outperforms the baselines on the real data (Table \ref{tab:real_data_results}), the $\mu$-F1 scores do not reach the high levels ($>0.90$) achieved on the synthetic data (Table \ref{tab:length_effect}). We partly attribute this performance gap to some noise (symbols which are not events) that may be introduced during data discretization. In the synthetic data, each symbol represents a perfectly observed, deterministic discrete event \cite{AGDES}. In contrast, the discretization technique for real data is based on the assumption of events as symbol/value-transitions (Section \ref{subsec:real_data set}). This transformation is inherently lossy where small changes in the measured continuous variable can generate fake state changes, while subtle but important events might be entirely overlooked. Consequently, this noise obscures the underlying structural dependencies between the cause and effect sequences, making it harder for our method to identify causal patterns through in-context learning.

\textcolor{blue}{Event symbols in the sequences carry no semantic meaning in our experiments and so the evaluation measures structural sequence information independently of semantic symbol meanings. This does not imply that semantic metadata should be ignored when it is available. Event names, system topology, or domain knowledge may provide additional evidence, and combining such information with our methodology remains outside the present evaluation.
}

\section{Acknowledgement}
This work was granted access to the HPC
resources of IDRIS under the allocation 2026-AD011016694 made by
GENCI.
\label{acknowledgement}

\bibliographystyle{ACM-Reference-Format}
\bibliography{BIB_MAIN_PAPER}

\end{document}